%% file: neurips_2025.tex
\documentclass{article}

\usepackage[preprint]{neurips_2025}
\input{math_commands.tex}

\usepackage[utf8]{inputenc} 
\usepackage[T1]{fontenc}    
\usepackage{url}            
\usepackage{booktabs}       
\usepackage{amsfonts}       
\usepackage{nicefrac}       
\usepackage{microtype}      
\usepackage{xcolor}         

\usepackage[dvipsnames]{xcolor}         
\definecolor{navy}{rgb}{0.0, 0.0, 0.5}
\usepackage[backref]{hyperref}
\usepackage{subcaption}
\usepackage{amsmath}
\usepackage{nicefrac}
\usepackage{todonotes}
\usepackage{changepage} 
\usepackage{tcolorbox}

\usepackage{algorithm}
\usepackage{algpseudocode}

\usepackage{pifont}
\newcommand{\cmark}{\ding{51}}%
\newcommand{\xmark}{\ding{55}}%

\newcommand{\fp}{\mathbf{h}} 
\newcommand{\ii}{x} 
\newcommand{\hidden}[1]{\mathbf{h}^{(#1)}} 
 \newcommand{\atomnum}{\mathbf{z}}
 \newcommand{\atompos}{\mathbf{r}}
\newcommand{\embed}{\tilde{x}}
\newcommand{\twonorm}[1]{\lVert #1 \rVert}
\newcommand{\embblock}{\mathcal{U}}
\newcommand{\outblock}{\mathcal{D}} 
\newcommand{\equiformer}{EquiformerV2}
\newcommand{\deq}{DEQuiformer}
\usepackage[normalem]{ulem}

\usepackage{array}
\newlength\savewidth
\newcommand{\tablestyle}[2]{\setlength{\tabcolsep}{#1}\renewcommand{\arraystretch}{#2}\centering\footnotesize}
\definecolor{baselinecolor}{gray}{.9}

\newcolumntype{x}[1]{>{\centering\arraybackslash}p{#1pt}}
\newcolumntype{y}[1]{>{\raggedright\arraybackslash}p{#1pt}}
\newcolumntype{z}[1]{>{\raggedleft\arraybackslash}p{#1pt}}

\newcommand*\circled[2][1.6]{\tikz[baseline=(char.base)]{
    \node[shape=circle, draw, inner sep=1pt,
        minimum height={\f@size*#1},] (char) {\vphantom{WAH1g}#2};}}

\title{DEQuify your force field: More efficient simulations using deep equilibrium models}

\author{%
  Andreas Burger\thanks{\texttt{andreas.burger@mail.utoronto.ca}} \\
  University of Toronto\\
  Vector Institute \\
  \And
  Luca Thiede\thanks{\texttt{luca.thiede@mail.utoronto.ca}} \\
  University of Toronto\\
  Vector Institute \\
  \AND
  Alán Aspuru-Guzik \\
  Acceleration Consortium \\
  University of Toronto\\
  Vector Institute \\
  NVIDIA \\
  \And
  Nandita Vijaykumar \\
  University of Toronto\\
}

\begin{document}

\maketitle

\begin{abstract}
%
Machine learning force fields show great promise in enabling more accurate molecular dynamics simulations compared to manually derived ones. 
Much of the progress in recent years was driven by exploiting prior knowledge about physical systems, in particular symmetries under rotation, translation, and reflections. In this paper, we argue that there is another important piece of prior information that, thus fa,r hasn't been explored: Simulating a molecular system is necessarily continuous, and successive states are therefore extremely similar. Our contribution is to show that we can exploit this information by recasting a state-of-the-art equivariant base model as a deep equilibrium model. This allows us to recycle intermediate neural network features from previous time steps, enabling us to improve both accuracy and speed by $10\%-20\%$ on the MD17, MD22, and OC20 200k datasets, compared to the non-DEQ base model. The training is also much more memory efficient, allowing us to train more expressive models on larger systems.
\end{abstract}

\section{Introduction}
With increasingly more available compute, molecular dynamics (MD) simulations emerged as an integral tool for studying the behaviour of molecules to develop a mechanistic understanding of a large class of processes in drug discovery and molecular biology \cite{lin2019force, hollingsworth2018molecular, sinha2022applications, durrant2011molecular}. 
The backbone of an MD simulation is a force field, which predicts the forces acting on each of the atoms in a molecule, given the current atom positions. These forces are then used to integrate the equations of motion numerically by multiplying the forces with a small time step $dt$ to obtain velocities, which in turn is used to update the atom's positions.
Traditionally, force fields were designed by hand to capture known physical effects such as covalent bonds, electrostatics, and van der Waals forces \cite{weiner1981amber, pearlman1995amber}. These hand-crafted force fields are compact and fast but lack the expressivity to capture more complex quantum mechanical many-body interactions. Alternatively, force fields can be calculated from highly accurate but costly quantum mechanical calculations, so-called ab-initio molecular dynamics (AIMD). Therefore, a new approach has gained traction over recent years: Training an expressive machine learning model on data from expensive ab-initio methods. This results in models at near-quantum chemical accuracy at only a fraction of the cost. 
\\
Some early works on machine learning force fields used local atom environment descriptors in combination with linear regression \cite{thompson2015spectral, shapeev2016moment}, Gaussian processes \cite{bartok2010gaussian}, and feed-forward neural networks \cite{behler2007generalized}. 
The pioneering work SchNet \cite{schutt2017schnet} used a rotation invariant graph neural network to predict energies, forces and other properties. This was later improved by the use of equivariant neural networks that model angular dependencies more directly, such as Cormorant \cite{anderson2019cormorant}, DimeNet \cite{gasteiger2020directional}, PaiNN \cite{schutt2021painn}, GemNet \cite{GemNet2021}, SphereNet \cite{liu2022spherical}, and NequIP \cite{NequIP2022}. Recent models have further improved the expressivity and scalability of equivariant models. Equiformer introduces an attention mechanism \cite{liaoEquiformerEquivariantGraph2023a}, Allegro focuses on edge features with non-growing receptive fields \cite{Allegro2023}, MACE introduces an efficient mechanism to calculate many-body interactions with high-order tensor polynomials \cite{MACE2022}, eSCN improves the scaling of Clebsch-Gordan products involved in equivariant convolutions \cite{eSCN2023}, and ViSNet \cite{wang2022visnet} and QuinNet \cite{wang2024efficiently} derive ways to incorporate four and five body terms much more efficiently. CHGNet \cite{Deng2023CHGNet} incorporates magmoms for extra physical supervision. VisNet-LSRM \cite{li2023long} and 4G-HDNNP \cite{Ko2021ChargeEquilibrium} focus on modeling long-range and non-local effects. 

Much of the architectural designs were guided by the incorporation of prior information about the systems, in particular, invariance to permutation of atom IDs, translation, rotation, and inversion symmetries of energy and forces, as well as size extensivity and smoothness under the movement of atoms. 
Our main observation is that in physically meaningful simulation, there is additional prior information that is not yet incorporated in any model: The evolution of atom coordinates in time needs to be continuous. In practical simulations, this is enforced by adaptively picking integration time steps small compared to the fastest moving part of the system. This way even processes that are seemingly abrupt on a macroscopic scale, such as shock simulations, are smooth on the atomic scale. Violation of this principle would lead to nonphysical energy dissipation. 
\\
We aim to use this extra information about our simulation by adapting the Deep Equilibrium Model (DEQ) framework \cite{DEQ2019} to equivariant architectures.  DEQs replace the typical deep stack of layers with a lightweight shallow model and a fixed-point solver, see section \ref{sec:deq}. This allows us to reuse latent features across simulation time steps by warm starting the fixed point solution from previous time steps, see \figref{fig:equiformer_vs_deq}. This lets us build effectively deep models at the cost of shallow ones.
Additionally, the formulation allows for much more memory-efficient training, permitting us to train expressive models on large systems that would otherwise not fit in memory. 
\\
Formulating energies and forces as fixed points is natural since the ab initio ground truth data used to train ML force fields are fixed points of self-consistent field (SCF) methods themself. This raises the hopes that our DEQ formulation induces an inductive bias leading to better generalization, and in section \ref{sec:results} we present some preliminary evidence for that. 
Tricks like restarting from previous fixed points are also used in ab-initio molecular dynamics simulations, where the density is initialized from previous time steps, or even more advanced extrapolation \cite{kolafa2004time} and Car-Parrinello schemes \cite{car1985unified}. Therefore, our formulation represents an interesting link between AIMD and ML MD.
\\
We implement our method by transforming the \equiformer{} architecture \cite{liaoEquiformerEquivariantGraph2023a, EquiformerV2024}, which has the highest accuracy on the Open Catalyst Project leaderboard \footnote{\href{https://opencatalystproject.org/leaderboard.html}{https://opencatalystproject.org/leaderboard.html}}, into a DEQ. In principle however, the methodology is compatible with other similar force field architectures.
Our results show that, compared to the original \equiformer{}, \deq{} achieves: 
(1) significantly improved accuracy for the OC20 200k dataset,
(2) 10-20\% faster inference and equally or higher accuracy on the MD17/MD22 datasets and in MD relaxations,  
(3) all at reduced training memory cost and (4) with up to 5x fewer model parameters.
\\
We summarize our contributions as follows:
\begin{enumerate}
    \item We identify the temporal continuity of molecular simulations as additional prior information that has not yet been used in any ML-based architecture. 
    \item We design the first DEQ-based equivariant neural network and apply it to ML force fields, which lets us exploit this temporal continuity by reusing fixed points across time steps 
    \item We demonstrate that our model can improve upon speed, accuracy, training memory, and parameter efficiency compared to non-DEQ counterparts on common benchmarks as well as real MD simulations
\end{enumerate}

\section{Background \label{sec:equivariantgnn}}
State-of-the-art ML force fields like \equiformer{} belong to the class of equivariant graph neural networks (GNN) \cite{MACE2022, Allegro2023, EquiformerV2024, NequIP2022}. 
The central shared feature is the stacking of equivariant message passing layers, typically between five \cite{NequIP2022} and twenty \cite{EquiformerV2024}.
3D rotational and translational equivariance is achieved by building on irreducible representations and spherical harmonics, improving data efficiency \cite{NequIP2022}. 
Such symmetry exploiting networks have emerged as the SOTA for molecular data \cite{eSCN2023, Allegro2023, MACE2022, EquiformerV2024, thomas2018tensor}. 
A Graph Neural Network (GNN) takes in a graph $\mathcal{G}$ and maps it to a target space in a permutation equivariant way. If the graph is embedded in 3d space as molecules are, we use $O(3)-$Equivariant graph neural networks, which are equivariant to translations, rotations and optionally inversions. 
In these networks, node features $h_t$ of node $t$, are concatenations of irreducible representations (irreps) $h^l_{t} \in \mathbb{R}^{2l + 1}$, organized by their degree $l$ (we omit an additional channel dimension for simplicity). Irreps transform under rotation $R$ as 
\begin{align}
    h^l(R \cdot (r_1,...,r_N)) = D_l(R) \cdot h^l(r_1,...,r_N)
\end{align}
where $r_1,...,r_N$ are the coordinates of the atoms, and $D_l(R) \in \mathbb{R}^{(2l+1) \times (2l+1)}$ is the Wigner-D matrix. Intuitively, higher-degree features rotate faster with rotation of the input features. $l=0$ features are rotation invariant scalars, and $l=1$ are ordinary vectors. A vector $r \in \mathbb{R}^3$ can be mapped to an $l$ graded feature using the spherical harmonics $Y_l(\nicefrac{r}{||r|}|) \in \mathbb{R}^{2l + 1}$. \\
Two irreps $f^{l_1}$ and $g^{l_2}$ of different degrees interact using the Clebsch-Gordan tensor product \cite{thomas2018tensor}
\begin{align}
    h^{l_3}_{m_3} = \left( f^{l_1}_{m_1} \otimes_{l_1, l_2}^{l_3} g^{l_2}_{m_2} \right)_{m_3} = \sum_{m_1 = -l_1}^{l_1} \sum_{m_2 = -l_2}^{l_2} C_{(l_1,m_1),(l_2,m_2)}^{(l_3,m_3)} f_{m_1}^{l_1} g_{m_2}^{l_2} \label{eq:tp}
\end{align}
where we index the elements within the $2l+1$ dimensional tensor by $m$, and $C_{(l_1,m_1),(l_2,m_2)}^{(l_3,m_3)}$ are the Clebsch-Gordan coefficients. Every combination of $l_1, l_2, l_3$ is called a path, and every path is weighted individually by $w_{l_1, l_2, l_3}(\cdot)$. The weight itself is predicted by a neural network, conditioned on rotation invariant features like the distance $||r_{ts}||$.\\
An equivariant GNN builts on top of \eqref{eq:tp} to define an equivariant message passing scheme: Given a target node $h_t$ and a source node $h_s$ with a relative coordinate vector $r_{ts}$, an equivariant GNN sends a message from the source to the target using
\begin{align}
    v_{ts}^{l_3} = v^{l_3}(h_t, h_s, r_{ts}) = \sum_{l_1, l_2} w_{l_1, l_2, l_3}(||r_{ts}||) \left( f^{l_1}(h_{t},h_{s}) \otimes_{l_1, l_2}^{l_3} Y^{l_2}(\nicefrac{r_{ts}}{||r_{ts}||}) \right) \label{eq:mp}
\end{align}
where $f(h_t, h_s)$ is a function of both target and source node features; in \equiformer{}, it is simply the concatenation operation. Instead of using \eqref{eq:mp} directly, \equiformer{} relies on eSCN convolutions, which calculates basically the same expression but in a more efficient way; please refer to \cite{eSCN2023} and \cite{EquiformerV2024} for details. 

\subsection{\equiformer{} \label{sec:equiformerv2}}

\begin{figure}
    \centering
    \includegraphics[width=\textwidth]{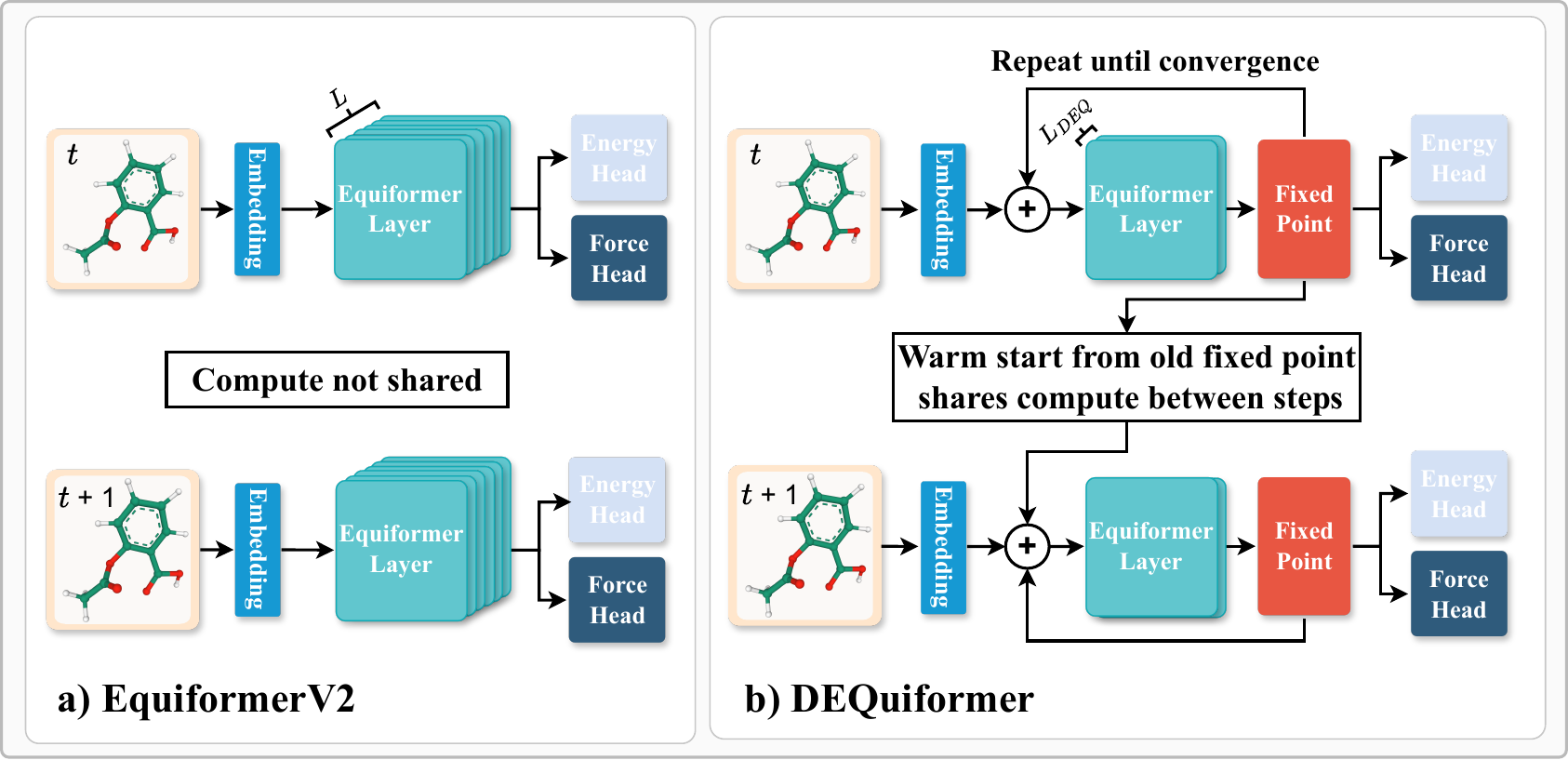}
    \vspace{1mm}
    \caption{Comparison of the \equiformer{} and \deq{} architectures. While the Equiformer model considers every input state independently, the \deq{} exploits the temporal continuity between input states to share compute. 
    This works because neighbouring time steps in an MD simulation are highly similar by design.
    Therefore, we drastically reduce the required compute by reusing the fixed-point from the previous step.}
    \label{fig:equiformer_vs_deq}
 \end{figure}

\equiformer{} \cite{EquiformerV2024} is a graph transformer, where each message passing layer is an equivariant transformer block. 
To initialize the node features, the embedding block $\embblock$ first encodes the input molecule, based the atom numbers $\atomnum$ and positions $\atompos$. 
\begin{align}
    \hidden{0}_i = \embblock(\ii_i) = \embblock\left(\atomnum_i, \{\atompos_{ij}\}_{j \in \mathcal{N}(i)}\right)
\end{align}
The $L$ transformer layers then perform repeated attention-weighted message passing to update the node features based on nodes in the neighborhood.
\begin{align}
    \hidden{l+1}_i = f^{(l)}_{\theta}\left(\hidden{l}_i, \{\mathbf{h}_j^{(l)}, \atompos_{ij}\}_{j \in \mathcal{N}(i)} \right)
\end{align}
After several transformer blocks update the node features, they are passed to two separate output heads for the final force and energy predictions. 
The total energy of the molecule is just the sum of the energies of the individual nodes.
\begin{align}
    E = \sum_i \outblock^{\text{scalar}} \left(\hidden{L}_i \right), \ 
    \mathbf{F}_i = \outblock^{\text{vector}} \left(\hidden{L}_i, \atomnum_i, \atompos_{ij} \right)_{j \in \mathcal{N}(i)}
\end{align}
We provide more details on equivariant GNNs and \equiformer{} in \secref{sec:equiformerarchitecture}.

\section{\deq{} \label{sec:dequiformer}}
Our goal is to incorporate temporal correlation as an inductive bias, to effectively reuse computation between timesteps of the simulation.
Inspired by SCF methods where the density can be initialized from previous calulations, we do so by reusing the features of previous timesteps.
It is unclear however how one could use latents from previous MD steps with an architecture like \equiformer{}.
The issue is that the model would have to learn to take into account previous latents during training. 
Fundamentally, there is no way for \equiformer{} to adapt to the (dis-)similarity from the previous timestep.
This is because explicit models define a fixed computational graph to map the input to an output. 
Instead, we use an implicit architecture like DEQ, that iterates until self-consistency is reached.
In this sense, DEQs (and other implicit models) have the ability to adaptively increase compute depending on how difficult the problem is. 
Concretely, we "DEQuify" \equiformer{} by replacing the $L$ Equiformer layers with a fixed-point solver over $L_{DEQ} \ll L$ Equiformer layers, as shown \figref{fig:equiformer_vs_deq}. In the following, we discuss the techniques to make this work in practice. In appendix \ref{sec:pseudocode} we provide pseudocode to summarize the DEQ algorithm and our modifications. 

\newcommand{\sstep}{s} 
\subsection{Deep Equilibrium Networks \label{sec:deq}}
\paragraph{Implicit layer}
Deep Equilibrium models \cite{bai2018trellis, DEQ2019} drastically reduce the model size by replacing the deep stack of layers with just one or two layers and an iterative fixed-point solver.
In \cite{DEQ2019}, the authors showed that the network converges to a fixed-point in the limit of infinite depth $L \rightarrow \infty$. 
Therefore, we say that these models have "continuous layers" or "infinite depth"
\\
To formalize this, consider a function $f_\theta$, usually a small neural network. 
Given some input $\ii$, repeated passes through $f_\theta$ updates the features $\fp^{\sstep}$ 
\begin{equation}
    \fp^{i+1} = f_{\theta}(\fp^{\sstep}, \ii)
\end{equation}
until the features converge to a fixed-point $\fp^{*}=f_{\theta}(\fp^{*}, \ii)$. The "fixed-point" or "equilibrium point" is then considered the output of the fixed-point layer.
This replaces the intermediate features $\hidden{l}$ after $l$ layers with a fixed-point estimate of the features $\fp^{\sstep}$. 
\equiformer{} predicts the forces and energy via separate output heads, by acting on the node features after $L$ layers $\hidden{L}$. 
The node features are instead replaced by the fixed-point estimate of the node features $\fp^{*}$ from the root solver, which we pass as input to the output heads.
\paragraph{Input injection via embedding block}
The neural network layer $f_{\theta}$ has to take in the input $\ii$, in addition to the current features $\fp^{\sstep}$, at every solver iteration, which is called the input injection.
Equiformer initializes the node features via an embedding block $\embed=\embblock\left( \ii \right)$. 
Using the embedding to initialize the initial fixed-point estimate $\fp^0$ however would stop gradients to flow to the encoder, since the gradient calculation is independent of the solver trajectory.
Instead, we use the embedding block's output as the input injection, by adding the embedding to the fixed-point estimate $\fp^{\sstep}$ at every solver step before passing it through the layer $g_{\theta}$.
The node features are instead initialized as all zeros $\fp^0 = 0$ \cite{DEQ2019}.
To prevent the norm of the features to grow with depth, we rescale the vector 2-norm $\twonorm{\cdot}$ to be the same as before the addition.
\begin{align}
    \label{eq:input_injection}
    f_{\theta}(\fp^{\sstep}, \ii) = g_\theta\left(
        (\fp^{\sstep} + \embed) \frac{\twonorm{\embed}}{\twonorm{\fp^{\sstep} + \embed}} 
    \right) 
\end{align}
This setup reduces the model size from many layers to just a few. However, naively passing $\fp$ through the NN layer $f_{\theta}$ until equilibrium is slow in practice.
Instead, we search for the fixed-point directly by using a root-solving algorithm, which computes more sophisticated updates of $\fp$ to reduce the number of passes until the fixed-point is reached. 
We found Broyden's method \cite{broydenClassMethodsSolving1965} to be instable during training, so we use Anderson acceleration \cite{anderson1965}.

\paragraph{Fast inference via fixed-point reuse}
The SCF-like structure of DEQs allows us to incorporate the inductive bias that consecutive time steps in a molecular dynamics simulation are highly similar. We do so by initializing the fixed-point estimate during inference not from all zeros, but the fixed-point of the previous time step: $\fp^{0}_{t+1}=\fp^{*}_{t}$ \cite{DEQOpticalFlow2022}.
With this fixed-point reuse the number of solver steps can be significantly reduced to gain a speedup at inference time. 

\paragraph{Memory efficient gradient}
Backpropagating through this solver trajectory would incur a prohibitive memory cost.
Fortunately, a unique feature of DEQs is that the gradient can be computed by the Implicit Function Theorem (IFT) \cite{DEQ2019}:
\begin{equation}
    \frac{ \partial L }{ \partial \theta } =
    \frac{ \partial L }{ \partial \fp^{*} }
    \left( 1 - \frac{ \partial f_{\theta} }{ \partial \fp^{*} } \right)^{-1}
    \frac{ \partial f_{\theta} \left( \fp^{*}, \ii \right) }{ \partial \theta } \label{eq:deq_grad}
\end{equation}
Using IFT, the forward passes is performed without tracking gradients, i.e. without storing the layer activations. 
Thus, the memory cost during training, becomes independent of the DEQ's "depth", which starkly contrasts explicit models, where the memory complexity grows linearly with each layer.
\\
With the Implicit Function Theorem (IFT) the gradient is computed by solving a second fixed-point system, for which we again use a root solver \cite{DEQ2019}.
\begin{equation}
\label{grad_fp_system}
\mathbf{g}^* = \mathbf{g}^* \frac{\partial f}{\partial \fp^*} + \frac{d L}{d \fp^*}
\end{equation}
Computing the gradient via IFT reduces the memory requirements during training, at the cost of extra time.
Recent DEQ works \cite{cao2024deqir, DEQOpticalFlow2022, geng2023diffusiondistillation} circumvent solving \eqref{grad_fp_system} by the so-called 1-step (phantom) gradient approximation \cite{fung2022jfb}. 
\begin{equation}    
\frac{ \partial L }{ \partial \theta } \approx  
    \frac{ \partial L }{ \partial \fp^{*} }
    \frac{ \partial f_{\theta} \left( \fp^{*}, \ii \right) }{ \partial \theta }
\end{equation}
We found however, that while the 1-step gradient leads to 2-3x faster training compared to solving the fixed-point system in \eqref{grad_fp_system}, it resulted in a significant reduction in accuracy \equiformer{}. 
We thus remove the 1-step gradient and use the IFT instead.

\paragraph{Recurrent dropout}
Dropout is a widely used regularization that tends to hurt DEQ performance.
This is because dropout samples a new mask for each pass through the implicit layer, which hinders finding a fixed-point \cite{DEQ2019}.
\equiformer{} uses two types of dropout, alpha dropout (acting on nodes) and path dropout, also known as stochastic depth (acting on edges). 
For \deq{} we instead use recurrent dropout, which applies the same mask at each step of the fixed-point solver, but a different mask for each sample \cite{DEQ2019,GalRecurrentDropout}.
We found that recurrent path dropout and no alpha dropout work best in \deq{}.

\paragraph{Training stability with fixed-point correction loss\label{sec:stability}}
Without further regularization, DEQs may become unstable over the course of training, noticeable by increasing number of root solver steps \cite{DEQOpticalFlow2022, baiStabilizingEquilibriumModels2021, gengTorchDEQLibraryDeep2023a}.
An effective remedy  is the \textit{sparse fixed-point correction} regularization loss \cite{DEQOpticalFlow2022}. 
Given a fixed-point solver trajectory 
$\fp^{0},\cdots,\fp^{\sstep},\cdots,\fp^{*}$ we pick some fixed-point estimates $\fp^{\sstep}, \sstep \in \mathcal{I}$ and add their gradient as if they were the final fixed-point estimate. We follow \cite{DEQOpticalFlow2022} and uniformly pick three $\fp^{\sstep}$ along the solver trajectory. 

\paragraph{Accuracy-compute tradeoff in the root solver \label{sec:solvertol}}
During training, we require low fixed-point errors to ensure that gradients can be calculated with the IFT. However, we can trade off performance and time during inference by relaxing the error threshold for the root solver \cite{DEQOpticalFlow2022}. 
With the right threshold, this significantly speeds up inference while only marginally affecting performance. 
For simplicity, we adhere to the settings of \cite{DEQOpticalFlow2022}.
During training we stop after the absolute fixed-point error falls below a relative threshold $|f_\theta\left(\fp^{\sstep}\right)-\fp^{\sstep}| / \twonorm{\fp^{\sstep}} < \epsilon_{train} = 10^{-4}$. During inference, we compute the first fixed-point at the same tight tolerance $\epsilon_{test} = \epsilon_{train}$, but then relax the threshold for the following time steps to $\epsilon^{FPreuse}_{test}=10^{-1}$. Relaxing the tolerance further reduces the number of forward steps and thus inference time, without sacrificing accuracy.

\section{Experiments}


\paragraph{OC20}
The Open Catalyst Project (OC20) is one of the largest quantum chemistry datasets, containing 1.3 million molecular relaxations from 260 million DFT calculations. It is focused on catalyst simulation, where each system consisting of a surface and an adsorbate that is relaxed onto the surface.
We train on the structure to energy and forces (S2EF) 200k split to evaluate the accuracy of our approach.
We then run relaxation simulations, starting from configurations in the dataset to validate that our fixed-point reuse scheme speeds up the simulation and does not induce additional errors.
The energies and forces are in units of eV and eV/\AA.
Time is measured as the forward pass on an AMD MI100.
\paragraph{MD17} 
MD17 contains one trajectory of a molecular dynamics simulations each for eight small molecules with 9 to 21 atoms. 
For each molecule, there are between 100,000 to 1,000,000 data points. 
Different to OC20, the data points in MD17 are from consecutive MD timesteps, which we will use to evaluate fixed-point reuse.
Following Equiformer, a random subset of 950 data points is used for training, 50 for validation, and testing on all remaining samples.
To aggregate the results over all molecules in MD17, we use minmax normalization per molecule and average over all molecules. We describe the procedure in \secref{sec:minmaxagg}. 
The energies and forces are in units of kcal/mol and kcal/mol/\AA.
\paragraph{MD22}
The MD22 dataset extends MD17 by seven larger molecules with 42 to 370 atoms \cite{md22_chmiela2023}. 
The size of the training set variies for each molecule. The number of samples is defined such that their sGDML model reaches a root mean squared test error for the forces of 1 kcal/mol/\AA. \cite{md22_chmiela2023}.
\paragraph{Training}
Following previous work, we train separate models for each molecule on MD17/22 and one big model on OC20.
We use the default model settings in the \equiformer{} repository for OC20 as well as MD17/22. The training hyperparameters were similarly taken from \equiformer{} for OC20, but slightly changed for MD17/22 to account for the smaller dataset sizes. We did not optimize the hyperparameters for \deq{}. Please refer to \secref{sec:hyperparameters} for details.

\subsection{Results}
\label{sec:results}
\begin{figure}
     \label{fig:error_vs_times_both}
     \centering
     \begin{subfigure}[t]{0.48\textwidth}
         \centering
         \includegraphics[width=\textwidth]{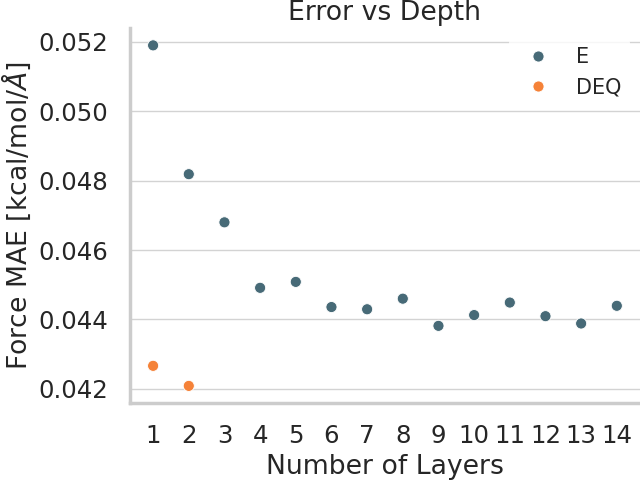}
         \caption{
         \textbf{Accuracy on OC20 200k}
         \deq{} outperforms \equiformer{} despite using much fewer parameters. Even with 14 layers (maxing out our memory), \deq{} still performs much better, indicating that DEQs are more data efficient for force fields.
         }
         \label{fig:error_vs_parameter}
     \end{subfigure}
     \hfill
     \begin{subfigure}[t]{0.48\textwidth}
         \centering
         \includegraphics[width=\textwidth]{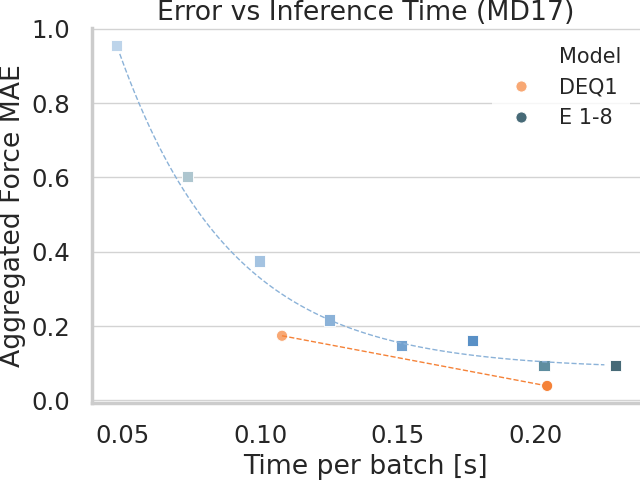}
         \caption{
         \textbf{Speed and accuracy on MD17}
         \deq{} is faster than Equiformer during inference at the same or better accuracy. Colors indicate number of layers from one to eight.
         }
         \label{fig:error_vs_time}
     \end{subfigure}
     \caption{Results on MD17/22 and OC20 200k: \deq{} is faster and more accurate than \equiformer{} while also using much fewer parameters.}
\end{figure}

\paragraph{Immproved accuracy on OC20}
We first demonstrate how DEQs improve peak accuracy by having expressivity comparable to very deep models.
To examine how the models scale with the number of layers, in \figref{fig:error_vs_parameter}, we plot the force error over the number of layers, using a maximum of 14 layers for \equiformer{}, the maximum our GPU memory could support, and up to two layers for our \deq{}. 
\deq{} reaches significantly better accuracy than \equiformer{} while using far fewer parameters.
Interestingly \equiformer{} seems not to benefit much from an increase in depth after a certain point, such that a one- or two-layer \deq{} is outperforming even an 14-layer \equiformer{}. This indicates that the formulation of force fields as DEQs might induce a useful inductive bias, due to the connection to SCF procedures. The results are in table \ref{tab:oc20_result}. To check for robustness of our results, we also train selected models for up to three times as long, with consistent results (see section \ref{sec:scale_compute}). 

\begin{figure}
     \centering
     \begin{subfigure}[t]{0.48\textwidth}
         \centering
         \includegraphics[width=\textwidth]{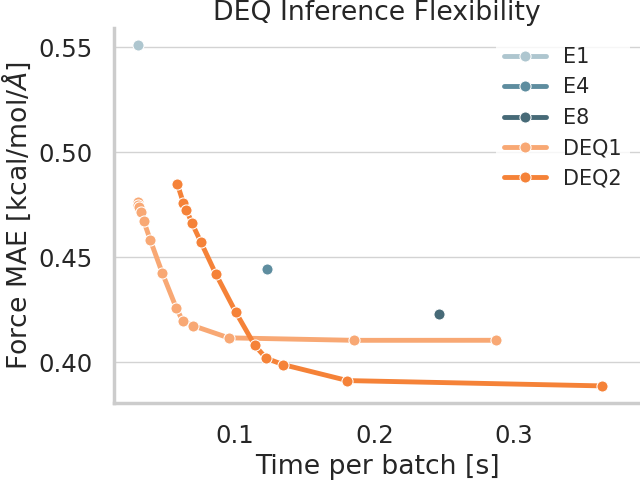}
         \caption{
         Compute-accuracy-tradeoff at inference time.
         \deq{} is remarkably robust to its fixed-point error up to a threshold of about $10^{-1}$, where the error starts to rapidly increase.
         As expected, higher fixed-point tolerances lead to faster inference speed.
         }
         \label{fig:acc_vs_time_fptolerance}
     \end{subfigure}
     \hfill
     \begin{subfigure}[t]{0.48\textwidth}
         \centering
         \includegraphics[width=\textwidth]{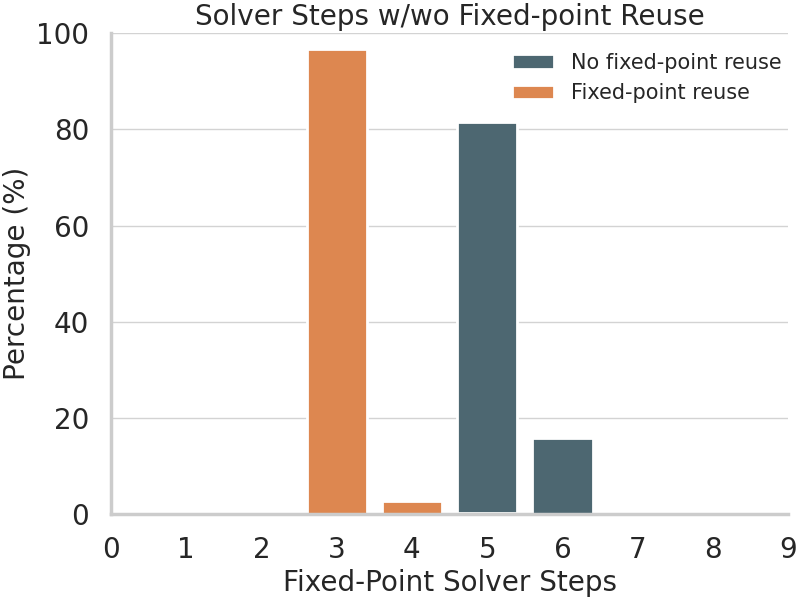}
         \caption{
         Reusing the fixed-point significantly reduces the number of solver steps in \deq{} to enable a speedup. 
         We plot two distributions, with and without fixed-point reuse.
         Percentage denotes the number of samples that required a given number of solver steps.
         }
         \label{fig:fpr_vs_fpnr}
     \end{subfigure}
     \caption{
     Reusing fixed-points and relaxing the solver threshold lead to better inference speed.
     }
\end{figure}

\paragraph{Speed-accuracy tradeoff at inference time}
While we get good accuracy with our models, a vanilla DEQ would be slow. To achieve a speedup we exploit the temporal continuity by reusing fixed-points across simulation time steps.
We examine the impact of fixed-point reuse in \figref{fig:fpr_vs_fpnr}. We plot the number of solver steps needed to find the fixed-point in the test set on Aspirin with and without reusing the fixed-point from the previous time step. Since MD17 contains one continuous MD trajectory per molecule, we can test on the whole trajectory and initialize the fixed-point of each sample with the fixed-point of the previous sample.
While \deq{} takes about 5-6 steps per sample to find the fixed-point when starting from zero initialization $\fp^0 = 0$; this gets reduced to 3 steps if we warm-start the solver, supporting our claim that we "share compute" between successive time steps by leveraging the temporal continuity.
\\
A unique feature of DEQs is that we can trade off accuracy for extra speed post-training by loosening the fixed-point error threshold. The looser this threshold, the faster the model, since the fixed-point solver terminates earlier. We are examining how sensitive the force error is with respect to this fixed-point error tolerance. We calculate the validation error and time per batch for different solver tolerances on a logarithmic scale, with the Aspirin molecule as an example. The results are plotted in \figref{fig:acc_vs_time_fptolerance}.
As expected, looser thresholds lead to faster inference time but higher force errors.
Remarkably, the model's predictions seem robust until a threshold of about $10^{-1}$, after which the force error shoots up. Thus, we choose a threshold of $10^{-1}$ as the sweetspot to further speedup inference.

\paragraph{Accuracy and speed on sequential MD data}
We also validate the accuracy and speed of \deq{} on the molecular dynamics data across MD17, using both fixed-point reuse and relaxed error threshold.  
For each molecule, we again test the inference speed on the full dataset, reusing the fixed-point from the previous sample to initialize the solver.
In \figref{fig:error_vs_time} \deq{} improves upon the Pareto front of various \equiformer{} layers. 
A full breakdown of the force test errors and inference time can be found in table~\ref{tab:md17_acc} and table~\ref{tab:md17_speed}. 
\deq{} achieves consistently faster inference speeds at the same or better accuracy than \equiformer{}. 
Comparing \deq{} to \equiformer{}, 
we measure an average inference time improvement of 19 \%, at 15 \% better accuracy. 
In total \deq{} is the best model in 5/8 molecules.
We also report the accuracy for the larger molecules in the MD22 dataset. \deq{} reaches state of the art accuracy in 3/7 cases and outperforms \equiformer{} on average.
The double-walled nanotube is much larger than the other systems, causing the 8-layer \equiformer{} to run out of memory on our compute setup while our \deq{} easily fits in memory. In appendix \ref{sec:scale_compute}, we also train models on Aspirin with up to 8 times more parameters to check that our results are robust across model sizes, with consistent results.

\begin{table*}[t]
\centering
\subfloat[
\textbf{Speed in relaxation simulation.} 
We ablate the impact of fixed point reuse and the lower error threshold $\epsilon^{FPreuse}_{test}$ (\secref{sec:solvertol}) on \deq{} and compare it to an \equiformer{}. Both techniques make a significant impact and \deq{} is faster than \equiformer{} when both are combined. 
\label{tab:relax_result}
]{
\centering
\begin{minipage}{0.53\linewidth}{\begin{center}
\tablestyle{4pt}{1.05}
\hspace*{-0.2cm}
\scalebox{0.70}{
\begin{tabular}{lllcc}
\toprule[1.2pt]
OC20 Relaxation & FP reuse & $\epsilon^{FPreuse}_{test}$ & Time [s] & \# Solver steps \\
\midrule[1.2pt]
\equiformer{} (14 layers) & & & $ 12.92 \pm 0.26  $ & -  \\
\midrule[0.6pt]
DEQ (1 layer) & \xmark & \xmark & $ 32.98 \pm 0.41  $ & $ 29.37 \pm 7.03 $ \\
DEQ (1 layer) & \cmark & \xmark & $ 20.37 \pm 0.43 $ & $ 18.05 \pm 2.19 $ \\
DEQ (1 layer) & \cmark & \cmark & $\mathbf{ 12.38 \pm 0.33 }$ & $ \mathbf{11.03 \pm 2.91} $ \\
\bottomrule[1.2pt]
\end{tabular}
}
\vspace{1.3mm} 
\end{center}}
\end{minipage}
}
\hspace{1.0em}
\subfloat[
\textbf{Accuracy on OC20.} 
The force and energy errors of various models on OC20 200k. \deq{} is more accurate than \equiformer{} on forces and energy.
Baselines from \cite{yang2024leignn}.
\label{tab:oc20_result}
]{
\centering
\begin{minipage}{0.4\linewidth}{\begin{center}
\tablestyle{4pt}{1.05}
\scalebox{0.73}{
\begin{tabular}{lccc}
\toprule[1.2pt]
OC20 200k [eV] [eV/\AA] & Force & Energy  & \# Weights \\
\midrule[1.2pt]
CGCNN & $0.075$ & $1.111$ & $3.6$M \\
SchNet & $0.060$ & $0.975$ & $7.4$M \\
MACE & $0.051$ & $0.565$ & $6.2$M \\
PaiNN & $0.053$ & $0.482$ & $13$M \\
PaiNN Direct & $0.047$ & $0.457$ & $14$M \\
DimeNet++ & $0.049$ & $0.497$ & $3.5$M \\
GemNet-dT & $0.041$ & $0.443$ & $32$M \\
LEIGNN & $0.044$ & $0.415$ & $17$M \\
\equiformer{} (8 layers) & $0.038$ & $\mathbf{0.392}$ & $3$M \\
DEQuiformer (2 layers) & $\mathbf{0.035}$ & $0.498$ & $\mathbf{1}$\textbf{M} \\
\bottomrule[1.2pt]
\end{tabular}
}
\vspace{1.5mm}
\end{center}}
\end{minipage}
}
\caption{
\deq{} is (a) faster in relaxation simulations (b) more accurate on OC20.
}
\label{tab:results_oc20_and_relax}
\end{table*}

\begin{table}[t]
\begin{adjustwidth}{-2.5cm}{-2.5cm} 
\centering
\scalebox{0.705}{
\begin{tabular}{lcccccccc}
\toprule[1.2pt]
MD17 Force MAE [kcal/mol/\AA] & \multicolumn{1}{c}{Aspirin} & \multicolumn{1}{c}{Benzene} & \multicolumn{1}{c}{Ethanol} & \multicolumn{1}{c}{Malonaldehyde} & \multicolumn{1}{c}{Naphthalene} & \multicolumn{1}{c}{Salicylic acid} & \multicolumn{1}{c}{Toluene} & \multicolumn{1}{c}{Uracil} \\
\cmidrule[0.6pt]{2-9}
\midrule[1.2pt]
DimeNet & $0.499$ & $0.187$ & $0.230$ & $0.383$ & $0.215$ & $0.374$ & $0.216$ & $0.300$ \\
PaiNN & $0.371$ & $230.000$ & $0.230$ & $0.319$ & $0.083$ & $\mathbf{0.209}$ & $0.102$ & $\mathbf{0.140}$ \\
SchNet & $1.350$ & $0.310$ & $0.390$ & $0.660$ & $0.580$ & $0.850$ & $0.570$ & $0.560$ \\
SphereNet & $0.430$ & $0.178$ & $0.208$ & $0.340$ & $0.340$ & $0.360$ & $0.155$ & $0.267$ \\
sGDML & $0.680$ & $\mathbf{0.060}$ & $0.330$ & $0.410$ & $0.110$ & $0.280$ & $0.140$ & $0.240$ \\
\equiformer{} (8 layers) & $0.359$ & $0.161$ & $0.175$ & $0.230$ & $\mathbf{0.063}$ & $0.243$ & $0.100$ & $0.219$ \\
DEQuiformer (2 layers) & $\mathbf{0.298}$ & $0.166$ & $\mathbf{0.162}$ & $\mathbf{0.216}$ & $\mathbf{0.063}$ & $0.218$ & $\mathbf{0.086}$ & $0.203$ \\
\bottomrule[1.2pt]
\end{tabular}
}
\end{adjustwidth} 
\begin{adjustwidth}{-2.5cm}{-2.5cm} 
\centering
\scalebox{0.75}{
\begin{tabular}{lccccccc}
\toprule[1.2pt]
MD22 Force MAE [kcal/mol/\AA]& \multicolumn{1}{c}{AT-AT} & \multicolumn{1}{c}{AT-AT-CG-CG} & \multicolumn{1}{c}{Ac-Ala3-NHMe} & \multicolumn{1}{c}{DHA} & \multicolumn{1}{c}{Stachyose} & \multicolumn{1}{c}{Buckyball} & \multicolumn{1}{c}{Nanotube} \\
\cmidrule[0.6pt]{2-8}
\midrule[1.2pt]
Allegro & $0.095$ & $0.128$ & $0.107$ & $0.073$ & $0.097$ & - & - \\
Frank & $0.099$ & $0.115$ & $0.088$ & $0.065$ & $0.088$ & - & - \\
GN-OC-L & $0.137$ & $0.130$ & $0.145$ & $0.091$ & $0.089$ & $0.189$ & $\mathbf{0.222}$ \\
GN-OC-S & $0.124$ & $0.134$ & $0.117$ & $0.066$ & $\mathbf{0.051}$ & $0.239$ & $0.258$ \\
GemNetOC & $0.124$ & $0.130$ & $0.117$ & $0.066$ & $0.089$ & - & - \\
Kovacs & $0.088$ & $0.106$ & $0.089$ & $0.053$ & $0.063$ & - & - \\
MACE & $0.099$ & $0.115$ & $0.088$ & $0.065$ & $0.088$ & $\mathbf{0.085}$ & $0.277$ \\
PaiNN & $0.238$ & $0.370$ & $0.230$ & $0.136$ & $0.233$ & - & - \\
SO3krates & $0.095$ & $0.128$ & $0.107$ & $0.073$ & $0.097$ & - & - \\
Shoghi & $0.216$ & $0.332$ & $0.244$ & $0.242$ & $0.435$ & - & - \\
TorchMD-NET & $0.204$ & $0.326$ & $0.188$ & $0.121$ & $0.192$ & - & - \\
sGDML & $0.691$ & $0.703$ & $0.797$ & $0.747$ & $0.674$ & - & - \\
\equiformer{} (8 layers) & $0.060$ & $0.039$ & $0.051$ & $\mathbf{0.043}$ & $0.092$ & $0.088$ & OOM \\
DEQuiformer (2 layers) & $\mathbf{0.059}$ & $\mathbf{0.038}$ & $\mathbf{0.050}$ & $0.044$ & $0.088$ & $0.092$ & $0.233$ \\
\bottomrule[1.2pt]
\end{tabular}
} 
\end{adjustwidth}
\vspace{2mm}
\caption{
\textbf{Accuracy on MD17 and MD22.} 
Force errors by molecular system in the MD17/22 dataset. Our \deq{} improves upon \equiformer{}'s accuracy on most systems.
The other baseline numbers are from \cite{schutt2021painn, gasteiger2020directional, liu2022spherical} (MD17) and \cite{li2024lrsm, shoghi2023jmp} (MD22).
}
\label{tab:md17_acc}
\end{table}

\paragraph{Markov property of DEQs}
We know from physics that that the forces should only depend on the current state, known as the Markov property.
By initializing our features from a previous time step, one might fear that this property is lost. To test if reusing fixed points breaks the Markov property, we compare the predicted forces with and without fixed-point reuse over Aspirin and OC20 relaxation trajectories. At each timestep, we calculate the relative difference in the forces as described in appendix \ref{sec:markov}. This results in an average deviation of less than $1\%$. Since the deviation is much smaller than the average prediction error, we conclude that fixed-point reuse approximately preserved the Markov property. This is congruent with results from ab initio molecular dynamics methods, where warm starting SCF iterations from density matrices of previous time steps do not affect the simulation, and are are therefore common practice.

\paragraph{Speedup in relaxations\label{sec:relax}}
Finally, we validate our findings in a real world test case.
To test the speedup of \deq{} in realistic simulation, we run relaxations based on configurations from OC20.
Each sample includes a slab model for the surface and an adsorbate on it as an initial guess. Starting from 100 samples of the OC20 200k data set we run 100 relaxation steps each to get the lowest energy geometry. 
We compare a one-layer \deq{} to the biggest model we can afford, the 14-layer \equiformer{}.
We also ablate reusing previous fixed-points and relaxing the solver threshold
and summarize the results in \ref{tab:relax_result}. 
We see that the speedup is only possible when using both techniques, reducing the number of layer evaluations significantly, from roughly 29 to 11. 
We find that \deq{} is faster than \equiformer{} in practical scenarios while being much smaller, and more accurate on the test set. 

\section{Limitations}
While we show that formalising ML-force fields as DEQ models lets us speed up computation by exploiting the temporal continuity of MD simulations, this speed up is still only modest compared to the billions of time steps necessary for many physically and biologically relevant processes. \\
Additionally, DEQs are slower to train, which may partially be offset by larger batch sizes due to reduced memory consumption, though.
\section{Conclusion}
In this work, we explored the integration of Deep Equilibrium models and machine learning force fields to enhance the efficiency of molecular dynamics (MD) simulations. 
We "DEQuify" the state-of-the-art model \equiformer{} by replacing its deep stack of layers with a more compact fixed-point layer. This approach allows us to leverage the temporal similarity between successive MD simulation states by reusing fixed-points, as well as the ability to trade off accuracy and speed.
On the MD17 and MD22 datasets, our \deq{} model achieves substantial improvements in parameter efficiency and inference speed at similar or better accuracy compared to the original \equiformer{}. 
On the much larger OC20 200k dataset, \deq{} reaches significantly higher accuracy compared to the base model.
This suggests a promising new research direction for machine learning force fields, focusing on exploiting the temporal nature of MD simulations to enhance computational efficiency. Since DEQs are in principal orthogonal to the base model, we expect that any improvements in the base architectures or DEQs in the future should complement each other. 
Future work could furhter expand on the idea of initializing features, akin to classical SCF methods like Pulay mixing, Anderson mixing, or density extrapolation methods.

\subsubsection*{Acknowledgments}
This research was undertaken thanks in part to funding provided to the University of Toronto’s Acceleration Consortium from the Canada First Research Excellence Fund CFREF-2022-00042.
Computations were performed on the Acceleration Consortium AMD Tacozoid cluster.
Resources used in preparing this research were provided, in part, by the Province of Ontario, the Government of Canada through CIFAR, and companies sponsoring the Vector Institute.
A.B. thanks the NSERC for the support through the Discovery Grant.

\newpage
{
    \bibliographystyle{plain}
    \bibliography{DEQ_bibtex}
}

\newpage
\appendix
\section{Appendix}

\subsection{Additional Background}




\subsubsection{\equiformer{} architecture \label{sec:equiformerarchitecture}}
We provide some further detail on the \equiformer{} architecture and its three main components: the embedding, the transformer layers, and the output heads. A complete description with additional details on layer norm, multi-head-attention and non-linearities can be found in the original papers \cite{liaoEquiformerEquivariantGraph2023a, EquiformerV2024}.  

\paragraph{Embedding} 
\newcommand{\atomindex}{i}
An input sample consists of the positions and types of all the atoms in the molecule.
The embedding block maps each atom $\atomindex$ to a higher dimensional node embedding $\hidden_\atomindex$, consisting of atom and edge-degree embeddings.
The edge-degree embeddings transform a constant one vector with an message passing SO(2) layer, multiplied with edge distance embeddings, and aggregated by summing.
Edge distance embeddings are the relative distances between the nodes, encoded by a learnable radial function on top of a Gaussian radial basis.
This sum is rescaled by a scalar $\alpha$ and added to an linear embedding of the one-hot atom number $\atomnum$:
\begin{align}
    u_t = \alpha \sum_{s \in \mathcal{N}(\atomindex)} v(1, 1, r_{\atomindex s})
\end{align}
\begin{align}
    \hidden_\atomindex = \embblock(\mathcal{G})_{\atomindex} = \text{linear}(\text{one-hot}(\atomnum_{\atomindex})) + u_{\atomindex}
\end{align}
$\mathcal{N}(t)$ means the neighbourhood of atom $\atomindex$, defined by the set of atoms that are within a user-specified cutoff radius from the atom $\atomindex$.

\paragraph{EquiformerBlock}
We write $\textit{EquiformerBlock}(\mathcal{G})$ to refer to a stack of $L$ Equiformer layers. Each layer consists of equivariant graph attention, layer norm and feed-forward networks. The equivariant graph attention updates the node features $h$ using equivariant messages (\eqref{eq:mp}). However, instead of just summing up the messages directly to update a target node, Equiformer weights each message with an attention weight to get the final message which is then summed over all source nodes:
\begin{align}
    m_{ts} &= a_{ts} \cdot v_{ts} \\
    h'_t &= h_t + \text{linear}\left(\sum_{s \in \mathcal{N}(t)} m_{ts} \right)
\end{align}
The attention weights are calculated using MLP attention \cite{liaoEquiformerEquivariantGraph2023a, brody2021attentive} operating only on the rotation invariant $L=0$ features:
\begin{align}
    z_{ts} = k^\top \text{LeakyReLU}(f(h^0_t, h^0_s))
\end{align}
\begin{align}
    a_{ts} = \frac{\exp{(z_{ts})}}{\sum_{k \in \mathcal{N}(t)} \exp{(z_{tk})}} 
\end{align}
with a learnable weight vector $k$. 

\paragraph{Output Heads}
The output heads take all the node features and process them depending on the type of target. For the energy, the $l=0$ features of each node are transformed by an MLP and summed together for the final prediction. For the forces, an additional layer of equivariant graph attention is used, and the $l=1$ features of each atom are directly treated as the prediction for the force.

\subsubsection{Inexact gradients in DEQ \label{sec:phantomgrad}}
The computational bottleneck in \eqref{eq:deq_grad} is to compute the inverse. Previous work has therefore explored approximating it via its Neumann series, sometimes called the phantom gradient \cite{fung2022jfb, geng2021training}. Often, keeping only the first term (the identity) is good enough, which leads to the so-called 1-step gradient
\begin{equation}
    \frac{ \partial L }{ \partial \theta } \approx  
    \frac{ \partial L }{ \partial \fp^{*} }
    \frac{ \partial f_{\theta} \left( \fp^{*}, \ii \right) }{ \partial \theta }
\end{equation}
The 1-step gradient can be implemented by simply passing the fixed-point through the implicit layer one additional time, this time with tracked gradients using autograd.
Many recent works have used the 1-step gradient with great success \cite{cao2024deqir, DEQOpticalFlow2022, geng2023diffusiondistillation}.
We found however that while the 1-step gradient leads to 2-3x faster training compared to solving the fixed-point system in \eqref{grad_fp_system}, it resulted in a significant reduction in accuracy, which is why we do not use the 1-step gradient in this paper. 


\subsection{Method}

\paragraph{Aggregated metric over MD17/MD22 \label{sec:minmaxagg}}
\newcommand{\mae}{\text{MAE}}
\newcommand{\normmae}{\text{NormMAE}} 
\newcommand{\avg}{\text{Avg}} 
\newcommand{\sem}{\text{Sem}} 
We use minmax normalization to rescale the errors of the different models on each molecule to $[0,1]$, where the models are \deq{} and \equiformer{} with various number of layers
$M \in \{DEQ1, DEQ2, E1, E4, E8\}$
. 
To get summary statistics per model, we then take the mean (Avg) 
over all normalized molecules.
\begin{align}
    \label{eq:minmax_norm}
    \normmae_{M}^{mol} = \frac{\mae_{M}^{mol} - min_{M}\left(\mae_{M}^{mol}\right)}{max_{M}\left(\mae_{M}^{mol}\right) - min_{M}\left(\mae_{M}^{mol}\right)} 
\end{align}
\begin{align}
\label{eq:avgsem}
    \avg_{M} = \frac{1}{N_{mol}} \sum_{mol} \normmae_{M}^{mol}
\end{align}

\paragraph{Hyperparameters for MD17/MD22 \label{sec:hyperparameters}}
To facilitate a fair and straightforward comparison, we follow the hyperparameters set out by EquiformerV1 \cite{liaoEquiformerEquivariantGraph2023a} and \equiformer{} \cite{EquiformerV2024}.
Since \equiformer{} did not evaluate on MD17/MD22, we refer to the EquiformerV1 \cite{liaoEquiformerEquivariantGraph2023a} codebase for training settings, which also provided the training loop for MD17/MD22 of our implementation.
To be economical with our GPU resources in terms of training and inference speed, we used a smaller maximum feature degree of $l=3$ (from previously $l=6$), which was also used in EquiformerV1. We observed that benefits from higher $l$ are neglectable on small datasets like MD17, as \cite{EquiformerV2024} also noted for the similarly sized QM9 dataset. 
We kept all parameters of the optimizer identical to EquiformerV1. 

\paragraph{Hyperparameters for OC20 S2EF 200k}
\equiformer{} provides hyperparameters for OC20 2M, which we take as a proxy for the OC20 200k split we train on.
The only changes made to the \equiformer{} model are (1) a reduction in the number of layers down from $12$ and (2) limiting the maximum spherical harmonics degree to $l=3$, since the 200k split is ten times smaller than the 2M split, and because it significantly increases the computational cost.
\equiformer{} made minor changes to the optimizer parameters compared to V1.
A full breakdown of hyperparameters is in table~\ref{appendix:tab:md17_hyperparameters}.

\begin{table}[t]
\centering
\scalebox{0.75}{
\begin{tabular}{lll}
\toprule[1.2pt]
Hyperparameters & MD17/MD22 & OC20 S2EF 200k \\
\midrule[1.2pt]
Optimizer & AdamW & \\
Learning rate scheduling & Cosine with linear warmup & \\
Warmup epochs & $10$ & 0.1 \\
Initial learning rate & $1 \times 10 ^{-6}$   & $4 \times 10 ^{-5}$  \\
Maximum learning rate & $5 \times 10 ^{-4}$  & $2 \times 10 ^{-4}$ \\
Minimum learning rate & $1 \times 10 ^{-6}$   & $2 \times 10 ^{-6}$  \\
Number of epochs & $1000$ & $6$ \\
Batch size & $4$ & \\
Force loss metric & L2 MAE & L2 MAE \\
Energy loss metric & L2 MAE & L1 MAE \\
Force loss weight $\lambda_{F}$ & $80$ & $100$ \\
Energy loss weight $\lambda_{E}$ & $1$ & $2$ \\
\midrule
Weight decay & $5 \times 10^{-3}$ & $10^{-3}$ \\ 
Gradient norm clipping & 1000 & 100 \\
Dropout rate (alpha dropout) & $0.1$ \equiformer{} / $0$ DEQ & \\ 
Stochastic depth (path dropout) & $0.05$ & \\ 
Cutoff radius (\AA) & $5.0$ & $12$ \\
Maximum number of neighbors & $500$ & $20$\\
Number of radial bases & $128$ & \\
Maximum degree $l_{max}$ & $3$ & \\
Maximum order $M_{max}$ & $2$ & \\
Grid resolution of point samples $R$ & $14$ & \\
Hidden dimension in feed forward networks $d_{ffn}$ & $128$ & \\
Dimension of hidden scalar features in radial functions $d_{edge}$ & $128$ &  \\
Embedding dimension (spherical channels) $d_{embed}$ & $128$ &  \\
$f_{ij}^{(L)}$ dimension $d_{attn\_hidden}$ & $64$ &  \\
Number of attention heads $h$ & $8$ & \\
$f_{ij}^{(0)}$ dimension $d_{attn\_alpha}$ & $64$ & \\
Value dimension $d_{attn\_value}$ & $16$ &  \\
\midrule
DEQ root solver & Anderson & \\
Maximum number of forward steps (stopping criterion) & $40$ & \\
Absolute error tolerance (stopping criterion) training $\epsilon_{train}$ & $10^{-4}$ & \\
Absolute error tolerance (stopping criterion) inference $\epsilon_{test}$ & $10^{-1}$ & \\
Fixed-point correction loss terms & $3$ & \\

\bottomrule[1.2pt]
\end{tabular}
}
\vspace{2mm}
\caption{
Hyperparameters for \equiformer{} and \deq{}. 
Training hyperparameters for MD17/MD22 are taken from the EquiformerV1 codebase. 
Model parameters are reduced to roughly a quarter to match the smaller MD17/MD22 benchmark.
For OC20 training and model settings are taken from the \equiformer{} repository.
}
\label{appendix:tab:md17_hyperparameters}
\end{table}

\paragraph{Implementation}
We use the model of \equiformer{} from \href{https://github.com/atomicarchitects/equiformer_v2/tree/fa32143fd43f609bfafb2513c1b8ca957553da5d}{commit fa32143}, which depends on open catalyst \href{https://github.com/FAIR-Chem/fairchem/tree/5a7738f9aa80b1a9a7e0ca15e33938b4d2557edd}{commit 5a7738f}. The open catalyst repository (now called FairChem) has since undergone significant changes.
For MD17/MD22 we modify the training loop from the code of EquiformerV1 \cite{liaoEquiformerEquivariantGraph2023a} from \href{https://github.com/atomicarchitects/equiformer/tree/b7e7a0df7df69bde35f593772093ecce3f6824a6}{commit b7e7a0d}.
The DEQ solver is adapted from the TorchDEQ library \cite{gengTorchDEQLibraryDeep2023a}. 
\\
For MD17/MD22 each model is trained on a single AMD MI100 GPU with 32GB GPU RAM for 1000 epochs, which takes 12 to 72 hours.
For OC20 200k training takes about 40 to 120 hours for six epochs.

\subsection{Additional results}
\begin{figure}
     \centering
     \begin{subfigure}[t]{0.32\textwidth}
         \centering
         \includegraphics[width=\textwidth]{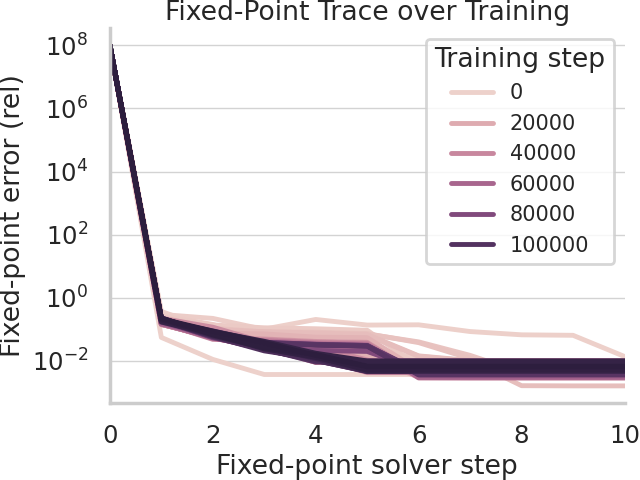}
         \caption{
         \deq{} converges stably to a fixed-point over training.
         }
         \label{fig:fpconvergence}
     \end{subfigure}
     \hfill
     \begin{subfigure}[t]{0.32\textwidth}
         \centering
         \includegraphics[width=\textwidth]{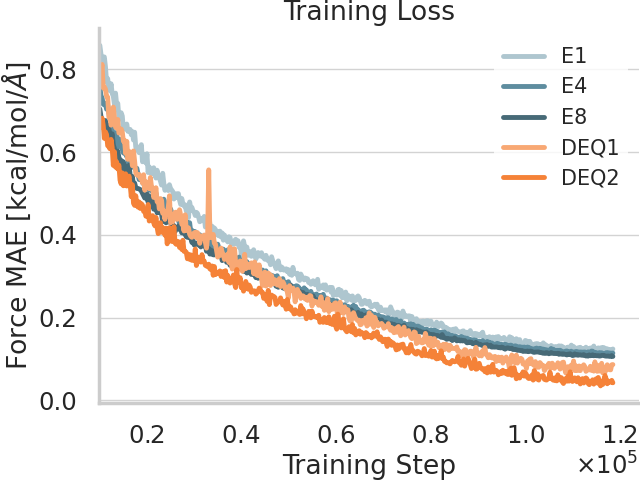}
         \caption{
         \deq{} trains faster, achieving lower train error.
         }
         \label{fig:trainerror}
     \end{subfigure}
     \hfill
     \begin{subfigure}[t]{0.32\textwidth}
         \centering
         \includegraphics[width=\textwidth]{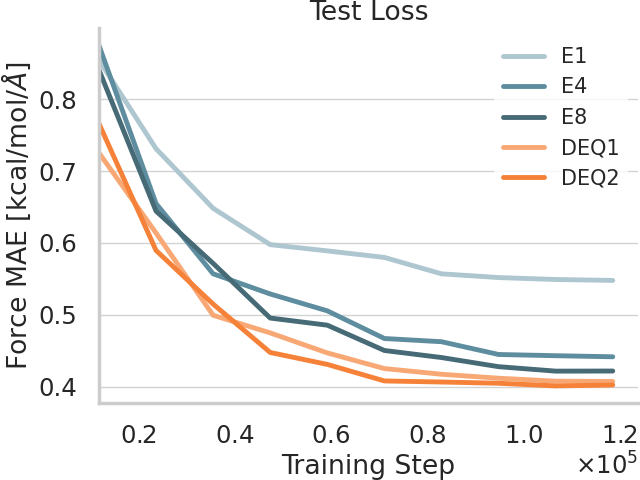}
         \caption{
         Lower train error translates to lower test error in \deq{}.
         }
         \label{fig:testerror}
     \end{subfigure}
     \caption{\deq{} enjoys stable training dynamics, reaching lower train and test error.}
     \label{fig:trainingrun}
\end{figure}

\paragraph{Training dynamics}
Our first question is whether or not \deq{} converges to a fixed-point. Since no prior work has combined DEQs with a rotation equivariant architecture, this is not at all obvious. To answer this question, we look at the relative fixed-point error on the Aspirin molecule as a function of the fixed-point solver steps at different steps in training; see \figref{fig:fpconvergence}. We see that the fixed-point error decreases with the number of solver steps, as expected. The fixed-point iteration is stable over the training, even slightly improving, resulting in slightly faster convergence later in training. \\
Additionally, we look at the loss curves of aspirin as an example, \figref{fig:trainerror} and \figref{fig:testerror}. We see that the training stability of \deq{} is similar to that of \equiformer{}.
The same holds for OC20 200k; see \figref{fig:trainerror_oc20} in the supplementary material.
We can also see that our \deq{} has both better training and better test errors, indicating that the improvements are not just due to reduced overfitting from a lower parameter count.

\paragraph{Training run on OC20}
In \figref{fig:trainingrun} of the main text we depicted that \deq{} achieves lower train and test error than \equiformer{} throughout training on Aspirin. 
For completion we also plot the training run for OC20 200k in \figref{fig:trainingrun_oc20}. Note that the choppy behavior of the training curve is due to resets of averaging statistics after each epoch.

\begin{figure}
     \centering
     \begin{subfigure}[t]{0.48\textwidth}
         \centering
         \includegraphics[width=\textwidth]{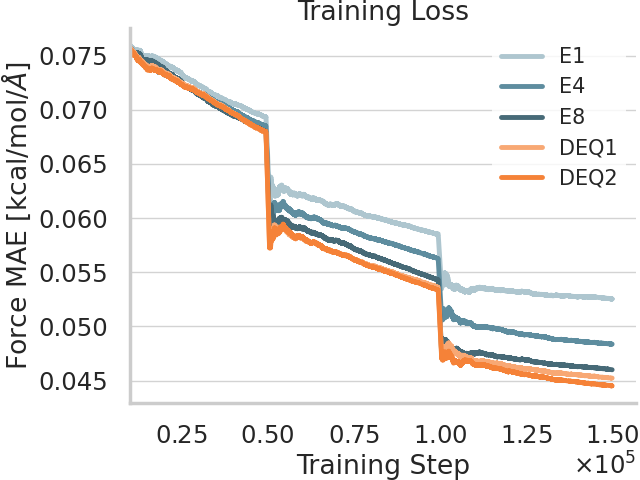}
         \caption{
         \deq{} trains faster, achieving lower train error. 
         We plot the error averaged over the current epoch. The step-like jumps are due to resetting the average at the start of a new epoch.
         }
         \label{fig:trainerror_oc20}
     \end{subfigure}
     \hfill
     \begin{subfigure}[t]{0.48\textwidth}
         \centering
         \includegraphics[width=\textwidth]{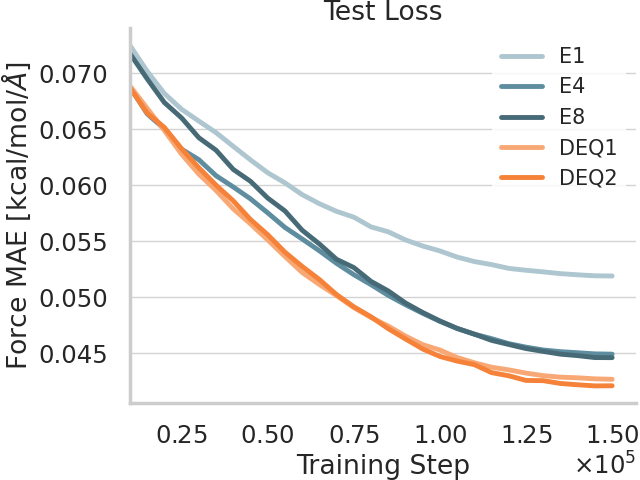}
         \caption{
         Lower train error translates to lower test error.
         }
         \label{fig:testerror_oc20}
     \end{subfigure}
     \caption{\deq{} enjoys stable training dynamics, reaching lower train and test error than \equiformer{} on OC20 200k.}
     \label{fig:trainingrun_oc20}
\end{figure}
\paragraph{MD-17 timings in detail}
A detailed table with runtimes per molecule in MD17 is given in table \ref{tab:md17_speed} 
\begin{table}[t]
\begin{adjustwidth}{-2.5cm}{-2.5cm} 
\centering
\scalebox{0.60}{
\begin{tabular}{lcccccccc}
\toprule[1.2pt]
MD17 Inference Time [s] & \multicolumn{1}{c}{Aspirin} & \multicolumn{1}{c}{Benzene} & \multicolumn{1}{c}{Ethanol} & \multicolumn{1}{c}{Malonaldehyde} & \multicolumn{1}{c}{Naphthalene} & \multicolumn{1}{c}{Salicylic acid} & \multicolumn{1}{c}{Toluene} & \multicolumn{1}{c}{Uracil} \\
\cmidrule[0.6pt]{2-9}
\midrule[1.2pt]
\equiformer{} (8 layers) & $0.218$ & $0.230$ & $0.245$ & $0.224$ & $\mathbf{0.200}$ & $0.218$ & $0.243$ & $0.220$ \\
DEQuiformer (2 layers) & $\mathbf{0.215}$ & $0.094$ & $\mathbf{0.218}$ & $\mathbf{0.249}$ & $0.199$ & $0.210$ & $\mathbf{0.201}$ & $0.094$ \\
\midrule[1.2pt]
\equiformer{} (4 layers) & $0.131$ & $0.146$ & $\mathbf{0.127}$ & $0.132$ & $\mathbf{0.113}$ & $0.125$ & $\mathbf{0.129}$ & $0.121$ \\
DEQuiformer (1 layer) & $\mathbf{0.074}$ & $\mathbf{0.104}$ & $0.097$ & $\mathbf{0.094}$ & $0.073$ & $\mathbf{0.098}$ & $0.077$ & $\mathbf{0.084}$ \\
\bottomrule[1.2pt]
\end{tabular}
}
\end{adjustwidth} 
\vspace{2mm}
\caption{
\textbf{Inference time on MD17.} \deq{} is faster at comparable accuracy. 
We highlight the lowest time per batch comparing \equiformer{} (4 layers) to \deq{} (1 layer), and \equiformer{} (8 layers) to \deq{} (2 layers), since they respectively closest in the speed-accuracy pareto front.
}
\label{tab:md17_speed}
\end{table}

\begin{figure}
     \centering
     \begin{subfigure}[t]{0.48\textwidth}
         \centering
         \includegraphics[width=\textwidth]{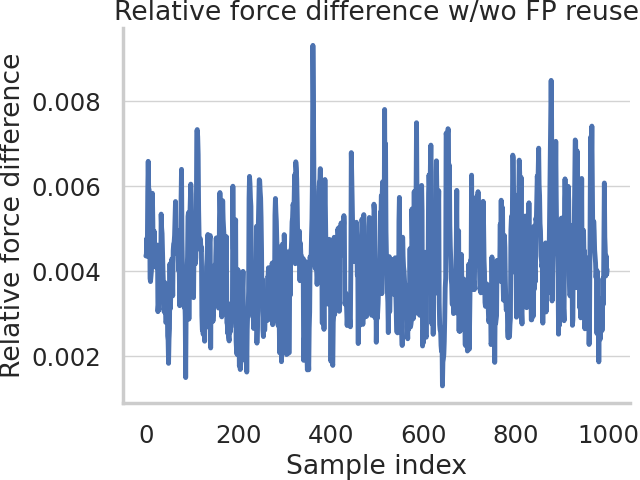}
     \end{subfigure}
     \caption{\textbf{Markov property.} 
     Initializing from the previous fixed-point, compared to initializing from zero, leads to very small deviation in forces $\Delta F_{rel}$ below one percent. This means, initialization from the past fixed point has almost no effect on the accuracy of the prediction.
     }
     \label{fig:forcereldiff_fpr}
\end{figure}

\paragraph{Fixed-point reuse approximately preserves Markovianity}
\label{sec:markov}
\newcommand{\forcevec}{\mathbf{F}}
An important property of molecular dynamics is that the forces only depend on the current state, known as the Markovian property.
To test if reusing fixed-points breaks Markov property, we compare the predicted forces $\forcevec$ with and without fixed-point reuse. At each timestep we calculate the relative difference in the forces as 
\begin{equation}
   \Delta F_{rel}^{atom}(atom\ i) = 
   \frac{
   |\forcevec^{fpr}_{i} - \forcevec_{i}|_{x}
   }{
   \frac{1}{2}\left(
    |\forcevec^{fpr}_{i}|_{x} + |\forcevec_{i}|_{x}
    \right)
   }
\end{equation} 
\begin{equation}
    \Delta F_{rel}^{sample}(sample\ j) = \frac{1}{N} \sum^{N}_{i \in atoms}\left( 
   \Delta F_{rel}^{atom}(i)
   \right)
\end{equation}
\begin{equation}
    \Delta F_{rel} = \frac{1}{M} \sum^{M}_{j \in test}\left( 
   \Delta F_{rel}^{sample}(j)
   \right)
\end{equation}
where $|\cdot|_{x}$ denotes the l2-norm over the three spatial components of a force vector on one atom $i$.
We run and average over $M=1$k consecutive samples of Aspirin from the MD17 dataset.
The relative force difference $\Delta F_{rel}$ is depicted in \figref{fig:forcereldiff_fpr}. We see a deviation in the predicted forces between starting from zero initialization and from the previous fixed-point of, on average, $0.4\%$. The deviation remains constant over time. We repeat the experiment for the 100 times 100 relaxation steps reported in \secref{sec:relax}, and measured a deviation of $0.8\%$. The deviation is much smaller than the average prediction error, so we conclude that fixed-point reuse approximately preserves the Markov property.

\begin{figure}
     \centering
     \begin{subfigure}[t]{0.48\textwidth}
         \centering
         \includegraphics[width=\textwidth]{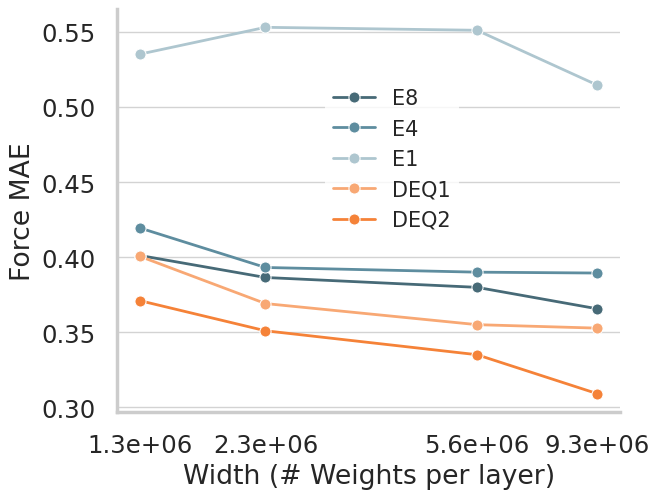}
         \caption{
         Error scaling with model width on Aspirin (MD17), trained at 1k epochs. On the x-axis is the number of parameters per layer. This means that the 8-layer Equiformer has approximately 8 times the parameter compared to the 1-layer DEQuiformer.
         }
         \label{fig:scale_aspirin}
     \end{subfigure}
     \hfill
     \begin{subfigure}[t]{0.48\textwidth}
         \centering
         \includegraphics[width=\textwidth]{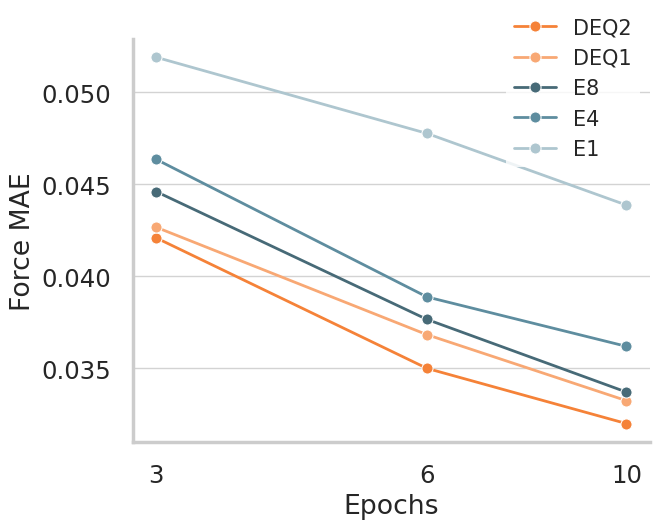}
         \caption{
         Error scaling with more epochs on OC20 200k. All models are getting better with more epochs, but DEQuiformer remains the leader in accuracy over Equiformer.
         }
         \label{fig:scale_oc20}
     \end{subfigure}
     \caption{
     Error scaling with more epochs and model size.
     }
     \label{fig:scaling_aspirin_oc20}
\end{figure}

\paragraph{Scaling compute}
\label{sec:scale_compute}
The paper directly compares \deq{} against \equiformer{}, albeit with limited compute compared to the \equiformer{} paper \cite{EquiformerV2024}, which trained up to 135M parameters on a larger datapslit (200k vs $>$100M) for $>$1500 GPU days.
\\
To demonstrate that our results are robust, we scale up selected runs.
In \figref{fig:scale_aspirin} we train Aspirin for the same number of epochs as in the paper (1000) at increasing model sizes. The smallest datapoint (left) is the same model size that we used in the main text for MD17/MD22, and the largest (right) the same as previously used for OC20. Note that at the same width \deq{} has much fewer total parameters, e.g. DEQ1$\sim$4.8M compared to E8$\sim$21M for the right-most width.
The accuracy gap between \deq{} to \equiformer{} remains when scaling the model size.
\\
We report the scaling with an increase in training epochs on OC20 200k in \figref{fig:scale_oc20}.
We did not scale up the model size, as \equiformer{} would run out of memory.
Instead we depict the same model size as in paper for OC20, with the left-most data point also trained on the same number of epochs. Again \deq{} increase in accuracy is robust when scaling up.


\paragraph{Pseudocode}
\label{sec:pseudocode}
To clarify our algorithm, we provide pseudocode for DEQuiformer in algorithm~\ref{alg:dequiformer} as well as for the original DEQ \cite{DEQ2019} in algorithm~\ref{alg:deq}.
\\
The original DEQ paper (algorithm~\ref{alg:deq}) is based on a transformer acting on a sequence of language tokens.
$x_{1:T}$ denote the input sequence and $y_{1:T}$ the output sequence of tokens.
$ f_{\theta}$ is a (weight tied) transformer layer.
\\
DEQuiformer (algorithm~\ref{alg:dequiformer}) acts on a cloud of atom positions and atom types. We drop the token indices $\cdot_{1:T}$ and omit the atom indices for readability. The \texttt{BackwardDEQ} procedure remains the same. The predicted and ground truth labels $y$ each consist of forces and the energy instead of sequences.
We made a couple of changes to the original DEQ. The original DEQ paper \cite{DEQ2019} used a linear initialization of the input injection, whereas we use EquiformerV2’s encoder. We also added a decoder (EquiformerV2's force and energy prediction heads). The solver is similar, but we use Anderson acceleration instead of Broyden’s method, where $\beta$ is the mixing parameter, $c_j$ are coefficients determined by minimizing the residuals, and
$ m $ is the number of previous iterations used in the mix \cite{anderson1965, gengTorchDEQLibraryDeep2023a}. We also add a normalization after each input injection. The original DEQ initialized fixed-points as zeros, whereas we took inspiration from \cite{DEQOpticalFlow2022} and initialized with the previous fixed-point during inference. From [\cite{DEQOpticalFlow2022} we also take the fixed-point correction loss and the relaxed solver tolerance $\epsilon$. The main change we made to EquiformerV2 was to remove alpha dropout as it hurt performance and replace path dropout with a recurrent path dropout (not shown in the algorithm). 

\newcommand{\zprev}{z^{*}_{t-1}}

\scalebox{0.7}{
\begin{minipage}{1.0\linewidth}
\begin{algorithm}[H]
\caption{Deep Equilibrium Model (DEQ), Bai 2019}
\label{alg:deq}
\begin{algorithmic}[1]
\Procedure{DEQ}{$\hat{x}_{1:T}, \theta, \epsilon$}
    \State Define $g_\theta(z_{1:T}; \hat{x}_{1:T}) = f_{\theta}(z_{1:T} + \hat{x}_{1:T}) - z_{1:T} $
    \State Initialize $z_{1:T}^{(0)} \gets 0$
    \State $i \gets 0$
    \While{$\|g_\theta(z_{1:T}^{(i)}; \hat{x}_{1:T})\| > \epsilon$} \Comment{fixed-point solver}
        \State $z_{1:T}^{(i+1)} \gets z_{1:T}^{(i)} - \alpha B g_\theta(z_{1:T}^{(i)}; \hat{x}_{1:T})$ \Comment{Broyden's method}
        \State $i \gets i + 1$
    \EndWhile
    \State $z_{1:T}^* \gets z_{1:T}^{(i)}$
    \State \Return $z_{1:T}^*$
\EndProcedure
\\
\Procedure{BackwardDEQ}{$z^*, y_{pred}, y_{gt}, \theta, \epsilon$}
    \State Compute $\frac{\partial \mathcal{L}}{\partial z^*}$ using the loss function $\mathcal{L}(y_{pred}, y_{gt})$
    \State Solve the linear system (IFT, second fixed-point solver):
    \[
    \left(J_{g_\theta}^\top \Big|_{z^*}\right) x + \left(\frac{\partial \mathcal{L}}{\partial z^*}\right)^\top = 0
    \]
    \State Compute the gradient:
    \[
    \frac{\partial \mathcal{L}}{\partial \theta} = -\left(\frac{\partial \mathcal{L}}{\partial z^*}\right) \left(J_{g_\theta}^{-1} \Big|_{z^*}\right) \frac{\partial f_\theta
    }{\partial \theta}
    \]
    \State \Return $\frac{\partial \mathcal{L}}{\partial \theta}$
\EndProcedure
\\
\Procedure{UseDEQ}{$x_{1:T}, y_{1:T}, \theta, \epsilon, \alpha$}
    \While{not done}
        \State $\hat{x}_{1:T} \gets W^T x_{1:T}$ \Comment{input injection}
        \State $z_{1:T}^* \gets \text{DEQ}(\hat{x}_{1:T}, \theta, \epsilon)$
        \State $y_{pred} \gets z_{1:T}^* $ \Comment{no decoder}
        \If{inference}
        \State $\frac{\partial \mathcal{L}}{\partial \theta} \gets \text{BackwardDEQ}(z_{1:T}^*, y_{pred}, y_{1:T}, \theta, \epsilon)$ 
        \State Update $\theta \gets \text{optimizer}(\theta, \frac{\partial \mathcal{L}}{\partial \theta})$ 
        \EndIf
    \EndWhile
    \State \Return $\theta$
\EndProcedure
\end{algorithmic}
\end{algorithm}
\end{minipage}
} 

\scalebox{0.7}{
\begin{minipage}{1.0\linewidth}
\begin{algorithm}[H]
\caption{DEQuiformer}
\label{alg:dequiformer}
\begin{algorithmic}[1]
\Procedure{DEQ}{$\hat{x}, \theta, \epsilon, \zprev$}
    \State Define $g_\theta(z; \hat{x}) = f_{\theta}\left((z + \hat{x}) \frac{||\hat{x}||}{||z+\hat{x}||} - z \right)$ \Comment{added normalization}
    \State Initialize $z^{(0)} \gets 0$ \Comment{if training}
    \If{inference}
        \State Initialize $z^{(0)} \gets \zprev$ \Comment{fixed-point reuse}
        \EndIf
    \State $i \gets 0$
    \State $\{z^{(i)}\} \gets \{\}$ \Comment{intermediate fixed-points for correction loss}
    \While{$\|g_\theta(z^{(i)}; \hat{x})\| > \epsilon$}
        \State $z^{(i+1)} \gets (1 - \beta) g(z^{(i)}; \hat{x}) + \beta \sum_{j=0}^m c_j z^{(i-j)}$ \Comment{Anderson acceleration}
        \If{training}
            \State $\{z^{(i)}\}$ append $z^{(i+1)}$ \Comment{if $i$ in $\mathcal{I}$, save intermediate fixed-point}
        \EndIf
        \State $i \gets i + 1$
    \EndWhile
    \State $z^* \gets z^{(i)}$
    \State \Return $z^*$, $\{z^{(i)}\}$
\EndProcedure
\\
\Procedure{UseDEQ}{$x, (\mathbf{F}_{gt}, E_{gt}), \theta, \epsilon, \alpha$}
    \State $\zprev \gets 0$ \Comment{if inference, save previous fixed-point}
    \While{not done}
        \State $\hat{x} \gets \text{Enc}(x) $ \Comment{input injection via Equiformer encoder}
        \State $z^*, \{z^{(i)}\} \gets \text{DEQ}(\hat{x}, \theta, \epsilon, \zprev)$
        \State $\zprev \gets z^*$ \Comment{save for fixed-point reuse}
        \State $\mathbf{F} \gets \text{Dec}_F(z) $ \Comment{Eqiformer decoder}
        \State $E \gets \text{Dec}_E(z) $ \Comment{Eqiformer decoder}
        \If{training}
            \State $\frac{\partial \mathcal{L}}{\partial \theta} \gets \text{BackwardDEQ}(z^{*}, (\mathbf{F},E), (\mathbf{F}_{gt}, E_{gt}), \theta, \epsilon)$ 
            \For{$z^{(i)}$ in $\{z^{(i)}\}$} \Comment{sparse fixed-point correction loss}
                \State $\frac{\partial \mathcal{L}}{\partial \theta} += \text{BackwardDEQ}(z^{(i)}, (\mathbf{F},E), (\mathbf{F}_{gt}, E_{gt}), \theta, \epsilon)$ 
            \EndFor
        \State Update $\theta \gets \text{optimizer}(\theta, \frac{\partial \mathcal{L}}{\partial \theta})$ 
        \EndIf
    \EndWhile
    \State \Return $\theta$
\EndProcedure
\end{algorithmic}
\end{algorithm}
\end{minipage}
}

\begin{figure}
     \centering
     \begin{subfigure}[t]{0.48\textwidth}
         \centering
         \includegraphics[width=\textwidth]{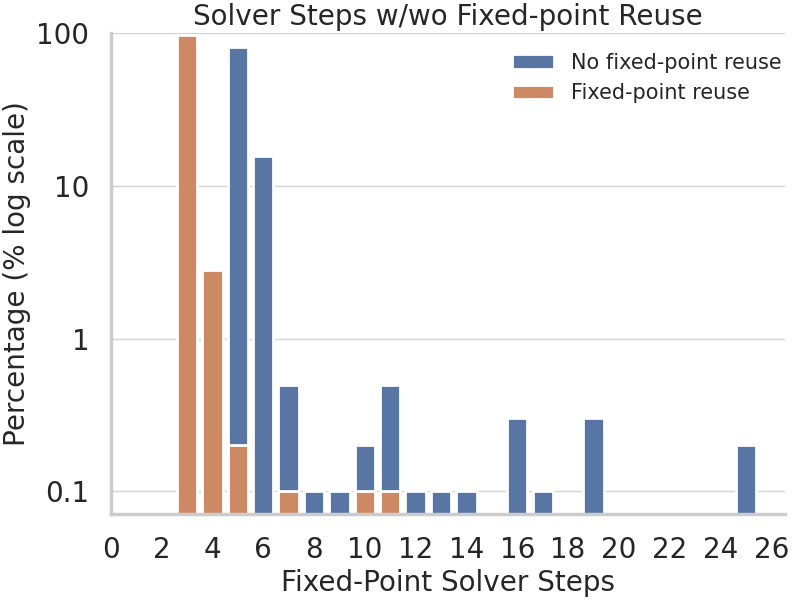}
         \caption{
         Reusing the fixed-point significantly reduces the number of solver steps in \deq{} to enable a speedup. 
         We plot two distributions, with and without fixed-point reuse.
         Percentage denotes relative number of samples in the test set that required a given number of solver steps.
         }
         \label{fig:fpr_vs_fpnr_log}
     \end{subfigure}
     \caption{
     Examining \deq{}s fixed-point behaviour.
     }
\end{figure}

\paragraph{Distribution over solver steps}
In \figref{fig:fpr_vs_fpnr} in the main text, the distribution over solver steps for without fixed-point reuse (blue) seemingly does not add up to 100\%.
We plot the same data again on a log scale on the y-axis in \figref{fig:fpr_vs_fpnr_log}.
There is a long tail of up to 25 solver steps for no fixed point reuse, that individually contribute less than 1\%, too small to show up in the plot of the main text.

\section{Broader impact} \label{sec:impact}
Molecular dynamics is a fundamental tool in computational chemistry, and by accelerating it our DEQ based architecture can advance scientific discovery. As is common with scientific application, the impact depends on how it’s applied. The biggest benefit could be seen drug‐discovery workflows.


\newpage
\section*{NeurIPS Paper Checklist}



\begin{enumerate}

\item {\bf Claims}
    \item[] Question: Do the main claims made in the abstract and introduction accurately reflect the paper's contributions and scope?
    \item[] Answer: \answerYes{} 
    \item[] Justification: We claim that we can exploit temporal continuity in MD simulations using DEQs to obtain expressive deep models at the cost of shallow models. We have several experiments using typical benchmarking datasets illustrating that this is possible.

\item {\bf Limitations}
    \item[] Question: Does the paper discuss the limitations of the work performed by the authors?
    \item[] Answer: \answerYes{} 
    \item[] Justification: We included a limitation section that contextualizes our speedups and talks about the need to further speed up ML force fields to bridge the gap to physically relevant time scales. We also talk about the slower training times.

\item {\bf Theory assumptions and proofs}
    \item[] Question: For each theoretical result, does the paper provide the full set of assumptions and a complete (and correct) proof?
    \item[] Answer: \answerNA{} 
    \item[] Justification: Our paper build on theoretical results from previous work and applies it to a new fields. As such, no new mathematical proofs are given.

    \item {\bf Experimental result reproducibility}
    \item[] Question: Does the paper fully disclose all the information needed to reproduce the main experimental results of the paper to the extent that it affects the main claims and/or conclusions of the paper (regardless of whether the code and data are provided or not)?
    \item[] Answer: \answerYes{} 
    \item[] Justification: We provide all the hyperparameters and architectural changes to \equiformer{}. All the data is publicly available.

\item {\bf Open access to data and code}
    \item[] Question: Does the paper provide open access to the data and code, with sufficient instructions to faithfully reproduce the main experimental results, as described in supplemental material?
    \item[] Answer: \answerNo{} 
    \item[] Justification: The data is publicly available standard datasets. We will make our code available upon acceptance.

\item {\bf Experimental setting/details}
    \item[] Question: Does the paper specify all the training and test details (e.g., data splits, hyperparameters, how they were chosen, type of optimizer, etc.) necessary to understand the results?
    \item[] Answer: \answerYes{} 
    \item[] Justification: We use standard data splits on established benchmarks and all the hyperaprameters are given.

\item {\bf Experiment statistical significance}
    \item[] Question: Does the paper report error bars suitably and correctly defined or other appropriate information about the statistical significance of the experiments?
    \item[] Answer: \answerNo{} 
    \item[] Justification: The experiments are to expensive to run them repeatedly for error bars. However, we do run scaling experiments, which show that our results are robust across different experimental settings.

\item {\bf Experiments compute resources}
    \item[] Question: For each experiment, does the paper provide sufficient information on the computer resources (type of compute workers, memory, time of execution) needed to reproduce the experiments?
    \item[] Answer: \answerYes{} 
    \item[] Justification: We provide our training hardware (MI100 GPU) as well as the training time.
    
\item {\bf Code of ethics}
    \item[] Question: Does the research conducted in the paper conform, in every respect, with the NeurIPS Code of Ethics \url{https://neurips.cc/public/EthicsGuidelines}?
    \item[] Answer: \answerYes{} 
    \item[] Justification: We do not see any violations with the code of ethics. 

\item {\bf Broader impacts}
    \item[] Question: Does the paper discuss both potential positive societal impacts and negative societal impacts of the work performed?
    \item[] Answer: \answerYes{} 
    \item[] Justification: We discuss the societal impacts in \ref{sec:impact}. We do not expect any other impact than what is common in scientific methods development.
    
\item {\bf Safeguards}
    \item[] Question: Does the paper describe safeguards that have been put in place for responsible release of data or models that have a high risk for misuse (e.g., pretrained language models, image generators, or scraped datasets)?
    \item[] Answer: \answerNA{} 
    \item[] Justification: ML force field models generally do not pose significant risks.

\item {\bf Licenses for existing assets}
    \item[] Question: Are the creators or original owners of assets (e.g., code, data, models), used in the paper, properly credited and are the license and terms of use explicitly mentioned and properly respected?
    \item[] Answer: \answerYes{} 
    \item[] Justification: We cite all prior work and original papers for the datasets we used. The only packages we do not cite are the standard libraries, e.g. numpy, matplotlib etc.

\item {\bf New assets}
    \item[] Question: Are new assets introduced in the paper well documented and is the documentation provided alongside the assets?
    \item[] Answer: \answerNA{} 
    \item[] Justification: We do not introduce any new datasets, but only use established benchmarks.

\item {\bf Crowdsourcing and research with human subjects}
    \item[] Question: For crowdsourcing experiments and research with human subjects, does the paper include the full text of instructions given to participants and screenshots, if applicable, as well as details about compensation (if any)? 
    \item[] Answer: \answerNA{}{} 
    \item[] Justification: No crowdsourcing or human subjects were involved.

\item {\bf Institutional review board (IRB) approvals or equivalent for research with human subjects}
    \item[] Question: Does the paper describe potential risks incurred by study participants, whether such risks were disclosed to the subjects, and whether Institutional Review Board (IRB) approvals (or an equivalent approval/review based on the requirements of your country or institution) were obtained?
    \item[] Answer: \answerNA{} 
    \item[] Justification: No human subjects were involved

\item {\bf Declaration of LLM usage}
    \item[] Question: Does the paper describe the usage of LLMs if it is an important, original, or non-standard component of the core methods in this research? Note that if the LLM is used only for writing, editing, or formatting purposes and does not impact the core methodology, scientific rigorousness, or originality of the research, declaration is not required.
    \item[] Answer: \answerNA{} 
    \item[] Justification: LLMs were not part of our model

\end{enumerate}

\end{document}

%% file: math_commands.tex
\usepackage{amsmath,amsfonts,bm}

\def\figref#1{figure~\ref{#1}}

\def\secref#1{section~\ref{#1}}

\def\eqref#1{equation~\ref{#1}}

\def\1{\bm{1}}

\DeclareMathAlphabet{\mathsfit}{\encodingdefault}{\sfdefault}{m}{sl}
\SetMathAlphabet{\mathsfit}{bold}{\encodingdefault}{\sfdefault}{bx}{n}

